\documentclass[letterpaper]{article}
\usepackage{times}
\usepackage{helvet}
\usepackage{courier}
\usepackage{url}
\usepackage{graphicx}
\graphicspath{{images/}}
\usepackage[margin=1in]{geometry}
\usepackage{amsmath}
\usepackage[backend=biber]{biblatex}
\usepackage{algorithm}
\usepackage[noend]{algorithmic}
\floatname{algorithm}{Algorithm}

\title{Interpretable Reinforcement Learning with Ensemble Methods}
\author{Alexander Brown and Marek Petrik \\
University of New Hampshire, 105 Main St, Durham, NH 03824 USA
\date{August 9, 2018}
}

\begin{document}

\maketitle

\begin{abstract}
We propose to use boosted regression trees as a way to compute human-interpretable solutions to reinforcement learning problems. Boosting combines several regression trees to improve their accuracy without significantly reducing their inherent interpretability. Prior work has focused independently on reinforcement learning and on interpretable machine learning, but there has been little progress in interpretable reinforcement learning. Our experimental results show that boosted regression trees compute solutions that are both interpretable and match the quality of leading reinforcement learning methods. 
\end{abstract}

\section{Introduction}

Reinforcement learning continues to break bounds on what we even thought possible, recently with AlphaGo's triumph over leading Go player Lee Sedol and with the further successes of AlphaGoZero, which surpassed AlphaGo learning only from self-play \cite{Silver2017}. While the performance of such systems is impressive and very useful, sometimes it is desirable to understand and interpret the actions of a reinforcement learning system, and machine learning systems in general. These circumstances are more common in high-pressure applications, such as healthcare, targeted advertising, or finance \cite{Dhurandhar2016}. 

For example, researchers at the University of Pittsburgh Medical Center trained a variety of machine learning models including neural networks and decision trees to predict whether pneumonia patients might develop severe complications. The neural networks performed the best on their testing data, but upon examination of the rules of the decision trees, the researchers found that the trees recommended sending pneumonia patients who had asthma directly home, despite the fact that asthma makes patients with pneumonia much more likely to suffer complications. Through further investigation they discovered the rule represented a trend in their data: the hospital had a policy to automatically send pneumonia patients with asthma to intensive care, and because this policy was so effective, those patients almost never developed complications. Without the interpretability from the decision trees, it might have been much harder to determine the source of the strange recommendations coming from the neural networks \cite{Bornstein2016}.

Interpretability varies between machine learning methods and is difficult to quantify, but decision trees are generally accepted as a good way to create an interpretable model \cite{Dhurandhar2016}. A decision tree represents a sequence of decisions resulting in a prediction. Small decision trees are easy to show graphically, which helps people interpret and digest them. Although small decision trees are not particularly powerful learners, several trees can be grouped together in an ensemble using techniques such as boosting and bagging in order to create a single more expressive model. No definitive measure of interpretability has been devised, but for decision trees one could count the number of nodes to compare interpretability between trees. With this measure one could also compare ensembles of trees, and as long as the trees are kept small, an ensemble of a few trees could be as interpretable as a single slightly deeper tree. Thus a reasonable ensemble could still be considered interpretable. A more detailed description of decision trees is found in \cite{Hastie2009}.

Of ensemble-building algorithms, gradient boosting is of particular interest to this work. Gradient boosting iteratively builds an ensemble of learners, such as decision trees, in such a way that each new learner attempts to correct the mistakes of its predecessors, optimizing some differentiable loss function.

We propose two new ways to compute interpretable solutions to reinforcement learning problems. Firstly, we show, in some benchmark environments, that policy data from a successful reinforcement learning agent (i.e. a neural network) can be used to train an ensemble of decision trees via gradient boosting, and that such an ensemble can match the performance of the original agent in the environment. Secondly, we analyze some possible ways of constructing an interpretable agent from scratch by doing policy gradient descent with decision trees, similar to gradient boosting. 

The rest of this work is organized as follows. Section 2 briefly discusses prior and related work. Section 3 introduces the general reinforcement learning problem and the benchmark environments used. Section 4 describes our use of ensemble methods to learn a policy from a reinforcement learning actor. Section 5 explains how we combine policy gradient descent and gradient boosting to build interpretable reinforcement learning systems. In section 6 we present the results of our experiments with those two techniques. Then, we discuss the results and propose future research steps in section 7.

\section{Related Work}
 
 Our work makes cursory use of neural networks and a reinforcement learning algorithm called SARSA, detailed descriptions of which can be found in \cite{Hastie2009} and \cite{Sutton2016} respectively.

A lot of reinforcement learning has historically been done with neural networks, but there has been work in using other tools from supervised learning in reinforcement learning algorithms. The work of Sridharan and Tesauro is among the first to successfully combine regression trees with Q-learning \cite{Sridharan2000}. The authors of \cite{Ernst2005} propose two new ensemble algorithms for use in tree-base batch mode reinforcement learning. The general integration of classifiers in reinforcement learning algorithms is demonstrated in \cite{Lagoudakis2003}, with the goal of leveraging advances in support vector machines to make reinforcement learning easier to apply ``out-of-the-box." An extensive analysis of classification-based reinforcement learning with policy iteration is given by \cite{Lazaric2016}, which explores a variant of classification in policy iteration that weights misclassifications by regret, or the difference between the value of the greedy action and that of the action chosen. 

 Our policy gradient boosting is a form of a policy gradient algorithm, which is a class of reinforcement learning techniques which attempt to learn a policy directly, rather than learning any kind of value function, by performing gradient ascent on a performance measure of the current policy. The general form of this kind of algorithm is described in \cite{Williams1992}, which refers to the class of algorithms as REINFORCE algorithms. These algorithms can be weak by themselves, and can oscillate or even diverge when searching for an optimal policy, so they are often used in combination with a value function or function approximation, forming an "actor-critic" pair \cite{Szepesvari2009}. Our algorithm uses a value function approximation, but purely for the purpose of reducing variance, as described in \cite{Sutton2016}.

An overview of interpretable machine learning can be found in \cite{Velido2012}. Petrik and Luss show that computing an optimal interpretable policy can be viewed as a MDP and that this task is NP hard \cite{Petrik2016}. 

\section{Problem Setting}

We consider reinforcement learning tasks, which typically consist of an agent interacting with some environment in a series of actions, observations, and rewards, formalized as a Markov Decision Process (MDP). At each time step the agent chooses an action $a$ from a set of legal actions $A$, which is passed to the environment, which returns a new state $s \in S$ and reward $r$ to the agent, where $S$ is the set of all states. The agent chooses actions according to some policy $\pi$, which is a (potentially stochastic) mapping from $S$ to $A$. We conduct our experiments in two benchmark environments: cart pole and mountain car. 

The cart pole problem \cite{Barto1983} has been a staple reinforcement learning benchmark for many years. The idea is simple: a cart which can move in one dimension carries an upright pole. The cart can be pushed either left or right, and the goal is to balance the pole for as long as possible. It is a good benchmark because it is easy to visualize, and has a relatively simple solution. The benchmark's difficulty can also easily be scaled upwards indefinitely, by stacking additional poles on the initial one \cite{Michie1968}. We deal with a version of the problem which terminates when the pole falls too far to be recovered, or when 200 timesteps have elapsed. 

The mountain car environment, as used in \cite{Boyan1995} to test control with continuous states, involves a car whose task it is to drive up a hill from in a valley. The car cannot make it up the hill simply by driving forward from the middle of the valley, but rather must reverse up the other side of the valley first to gain momentum. A reward of -1 per time step is given until the car reaches the goal state, at which point the reward is 0 and the episode terminates. The episode also terminates when 200 timesteps have elapsed, if the car has not managed to traverse the hill to the goal.

\section{Supervised Learning on Reinforcement Data}

In order to compute an interpretable solution to an MDP, we propose to record policy data from a well-performing agent in that MDP, and train an ensemble of CART \cite{Breiman1984} trees on that data using gradient boosting. This requires both a working solution to the MDP and some simulation of the MDP from which data can be gathered.

Gathering the data for this approach is straightforward: both the states that the working solution encounters and the actions it selects are recorded for a number of trials. Then, a new ensemble of trees is trained to predict the actions from the states. For some environments this may be equally straightforward, but for others it require some transformation of the state or action spaces to be more conducive to tree prediction. Finally, the ensemble is evaluated in the simulation, and the rewards compared to those of the initial solution.

Performing supervised learning on reinforcement data is advantageous because it has the potential to utilize state of the art reinforcement learning algorithms and techniques. It is insensitive to the composition of the initial agent, as it merely requires a black box from which policy data can be extracted. This lends flexibility to the idea, and hints at the possibility of a wide range of uses.

The primary drawback is that it requires the MDP be at least approximately solved for any hope of reasonable performance. This will be addressed in Section 5, which provides a ground-up interpretable solution. It should also be noted that insights from the interpretability of the trees are not guaranteed to apply to the initial solution, as the trees merely studied its behavior. This is unlikely to be a significant drawback, however, unless it is desired to specifically understand the initial solution, in which case the interpretability of the trees is a reasonable start anyway.

\section{Gradient Boosting for Reinforcement Learning}

Applying gradient boosting to policy iteration has an intuitive appeal. Gradient boosting \cite{Friedman2001} works by training a new tree to compensate for the failures of the previous ensemble, in the direction of the gradient. This seems to fit well into batch policy gradient descent by using the update step to train a new tree according to the gradient of the policy, which in turn depends on the policy parameterization \cite{Williams1992}.

We implement a REINFORCE \cite{Williams1992} algorithm using a policy parameterized by a softmax of numerical action preferences, as in \cite{Sutton2016}. REINFORCE seems to lend itself to our application; its simplicity and extensibility let us easily construct an algorithm with a solid backbone that uses decision trees in the inner loop. The softmax of the $i$th element in a vecor $x$ of length $n$ is given by $$ \textrm{softmax}(x_i) = \frac{\exp(x_i)}{\sum_{j = 1}^{n} \exp(x_j)},$$ and the softmax of a vector is the vector that results from applying the softmax to each element. We use an approximate value function $v$ as a baseline, in an attempt to reduce variance \cite{Sutton2016}. The general steps to the algorithm are shown in \textbf{Algorithm 1}.

\begin{algorithm}
	\caption{REINFORCE with Trees}
	\begin{algorithmic}
		\REQUIRE approximate state-value function $v$
		\STATE ENSEMBLE $\gets$ empty set of trees
		\WHILE {training ensemble}
			\STATE $L \gets$ empty list
			\FORALL {episodes in batch}
				\STATE reset environment
				\WHILE {timestep $i$ in episode}
					\STATE read state $s_i$ from environment				
   					\STATE $p \gets$ empty list
					\FORALL {action $a$}
						\STATE $p_a \gets \frac{1}{|A|}$ where $|A|$ is the number of actions
					\ENDFOR
					\FORALL {trees $t$ in ENSEMBLE}
						\STATE compute weight vector $w_{s}$ according to $t$
						\STATE $p \gets p + w_{s}$
					\ENDFOR 
					\STATE $p \gets$ softmax$(p)$
					\STATE randomly select action $a_i$ according to action probability vector $p$
					\STATE reward $r_i \gets$ take $a_i$ in environment
					\STATE $L \gets L$ append ($s_i$, $a_i$, reward $r_i$, $i$)
				\ENDWHILE
			\ENDFOR
			\FORALL {$s, a, r, i$ in $L$}
				\STATE compute discounted future reward $G_i$
				\STATE compute value of $s$:  $v(s)$
				\STATE replace current $r$ in $L$ by $G_i - v(s)$ (see \cite{Sutton2016})
			\ENDFOR
			\STATE update $v$
			\STATE train tree $t$ to predict $a$ from $s$ in $L$
			\STATE ENSEMBLE $\gets$ ENSEMBLE append $t$
		\ENDWHILE
	\end{algorithmic}
\end{algorithm}

This algorithm continues to grow in space and computation time as it runs, by nature of endlessly adding trees. In the case where it needs to train for a long time, boundless growth becomes problematic, so we also examine a version which is capped at a fixed number of trees. The goal is to take some information from older trees and then discard them, passing the information along to new trees which occupy the freed space. This recycling is achieved by training new trees not only on the data from the most recent batch of episodes, but also on predictions from a tree about to be recycled.

\section{Experimental Results}

\begin{figure}
	\centering
	\begin{minipage}{0.45\linewidth}
		\centering
		\includegraphics[width=0.95\linewidth]{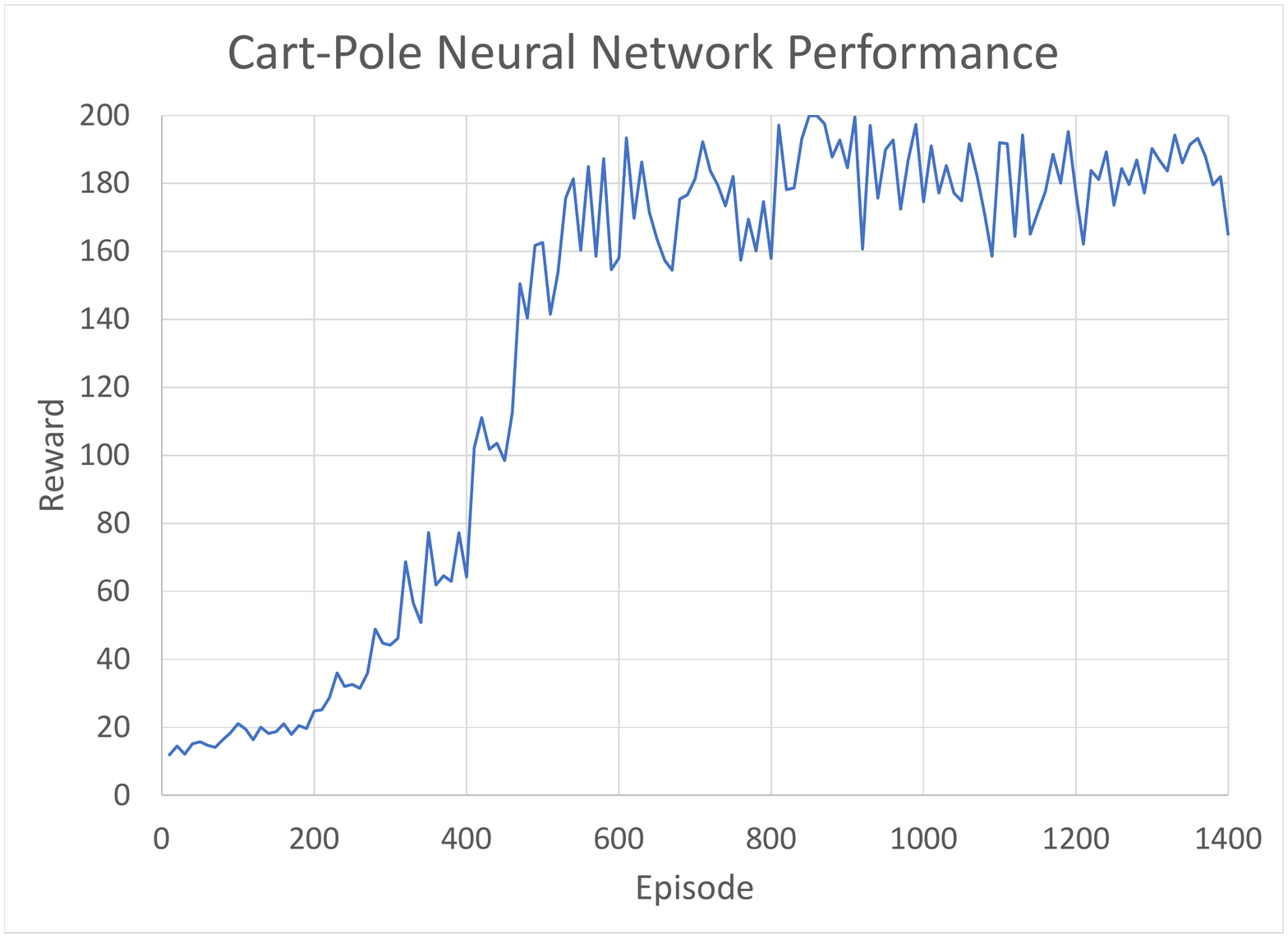}
		\caption{Neural Network training in the cart-pole environment}
		\label{fig:cart-pole-nn}
	\end{minipage}%
	\hfill
	\begin{minipage}{0.45\linewidth}
		\centering
		\includegraphics[width=0.95\linewidth]{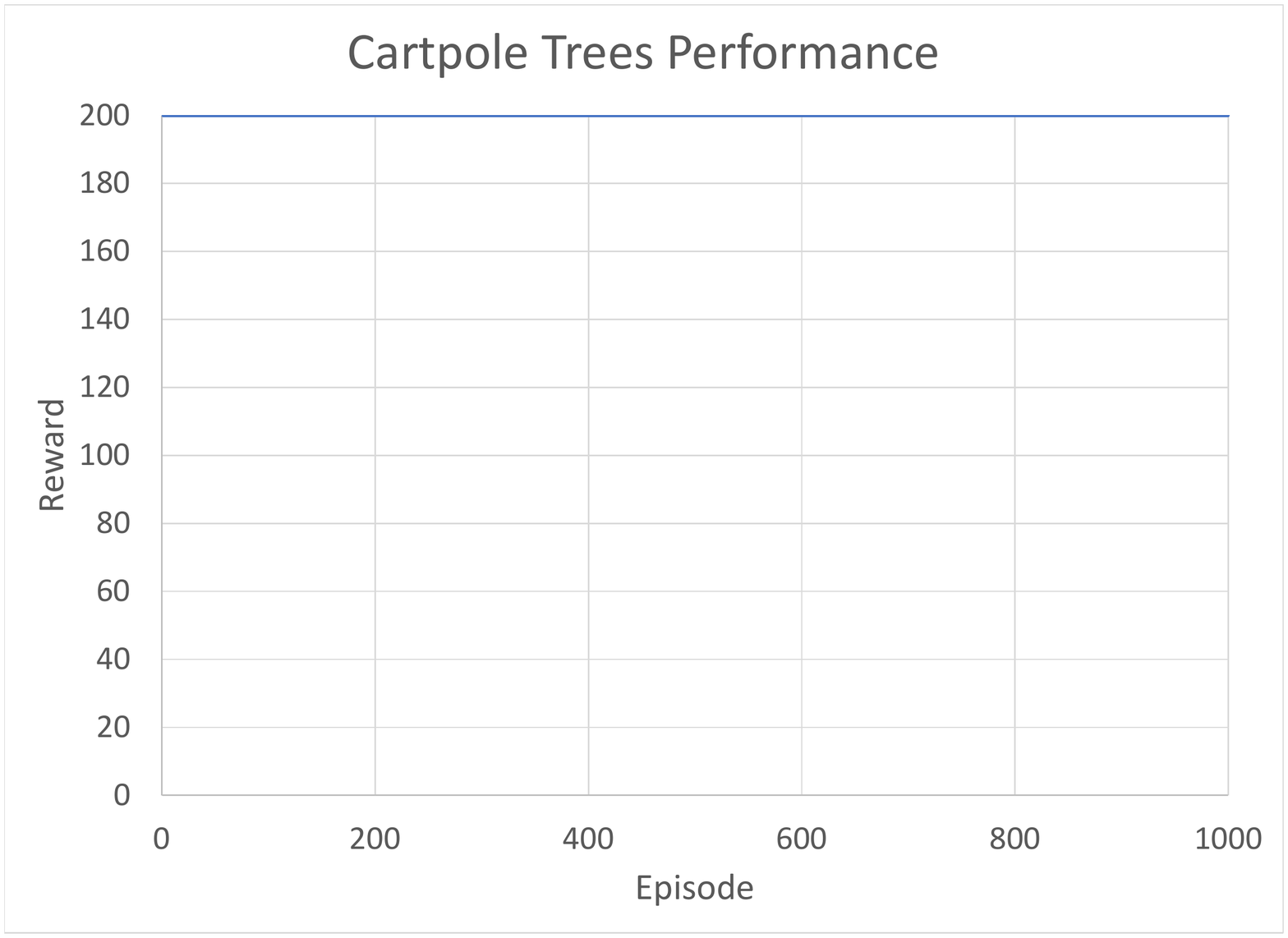}
		\caption{Ensemble trained from the neural network policy in the cart-pole environment}
		\label{fig:cart-pole-trees}
	\end{minipage}
\end{figure}

Supervised learning with reinforcement data was quite successful on both benchmark environments. In the cart-pole environment, the initial policy data was obtained from a neural network, the training of which is shown in Figure \ref{fig:cart-pole-nn}. As can be seen in the plot, the neural network's performance was less than perfect, which, although it indicates that the neural network could have been tuned to better suit the environment, serves to show something important about the results. The ensemble performed perfectly in the environment (see Figure \ref{fig:cart-pole-trees}), lasting the full 200 timesteps each episode, which is noticeably better than the data on which is was trained. We believe this to be a result of overfitting on the part of the neural network, and it is very promising that the ensemble was resistant enough to overfitting to outperform the neural network.

In the mountain car environment the initial policy data was obtained from a trained SARSA agent \cite{Sutton2016}. The ensemble did not outperform SARSA, as in the case of cart-pole, but it did match the episodic performance, which was the goal. This is also promising but not truly conclusive, as both environments are quite simple.

\begin{figure}
	\centering
	\begin{minipage}{0.45\linewidth}
		\centering
		\includegraphics[width=0.95\linewidth]{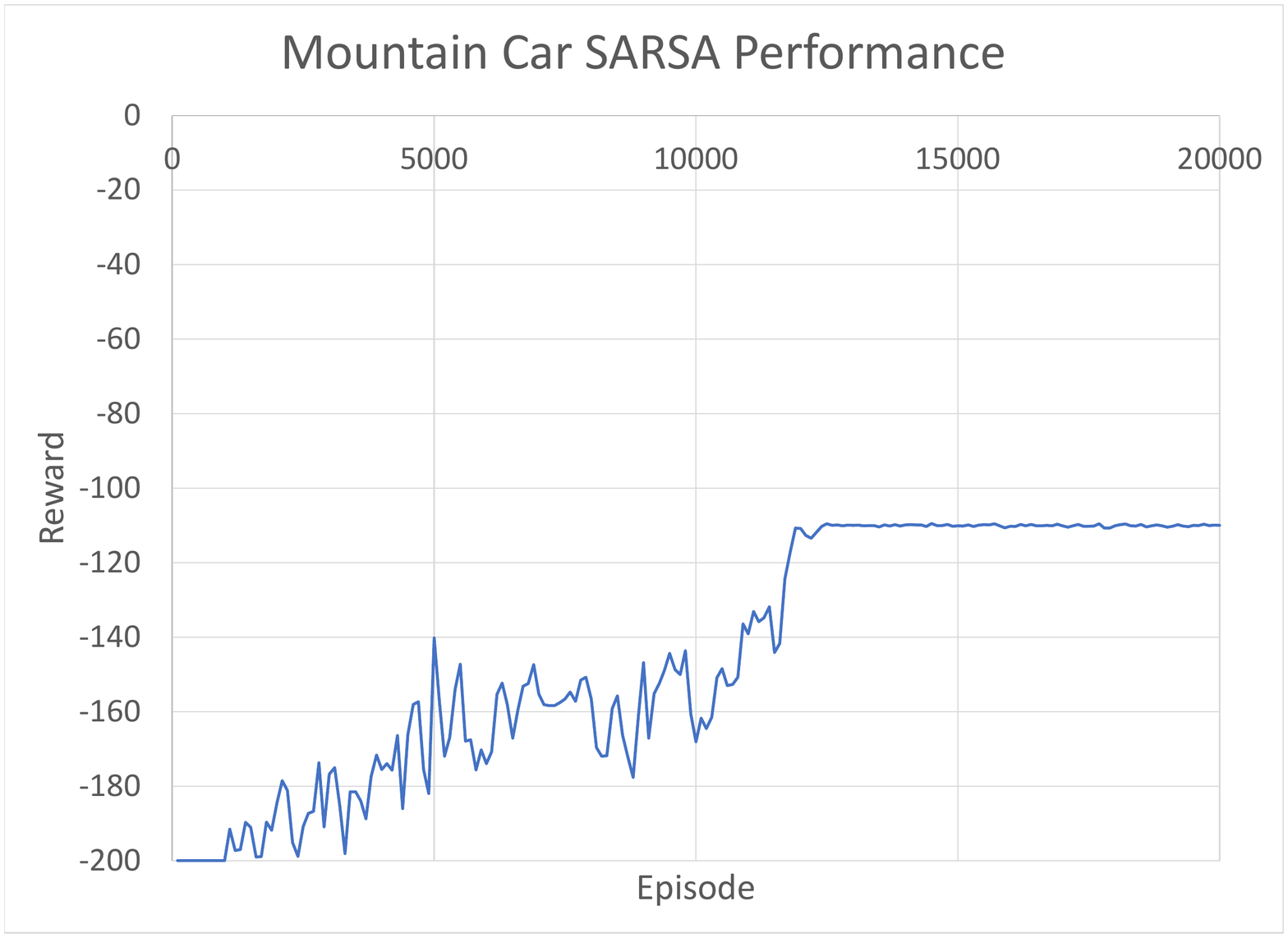}
		\caption{SARSA training in the mountain car environment}
		\label{fig:mountain-car-sarsa}
	\end{minipage}%
	\hfill
	\begin{minipage}{0.45\linewidth}
		\centering
		\includegraphics[width=0.95\linewidth]{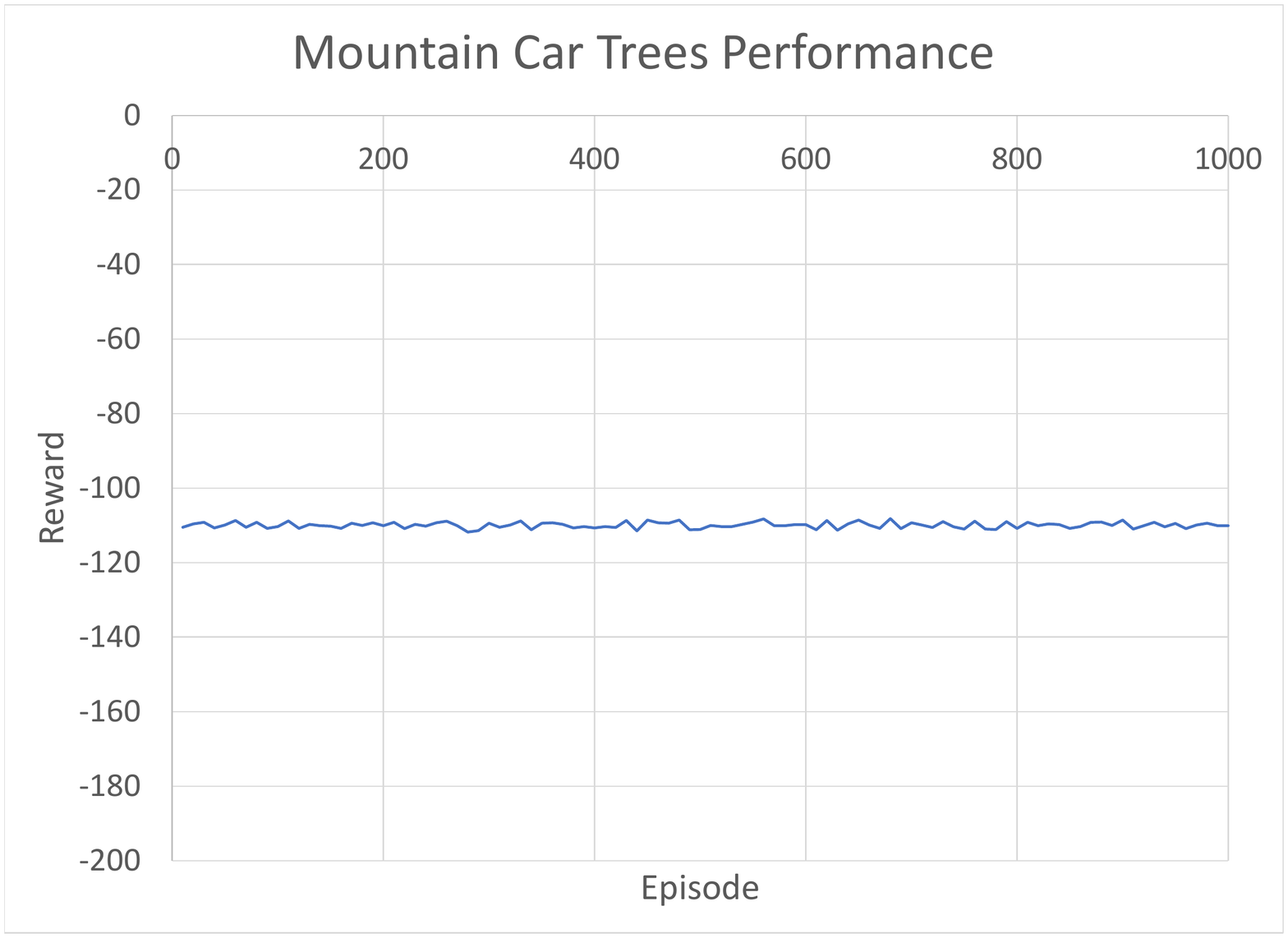}
		\caption{Ensemble trained from the SARSA policy in the mountain car environment}
		\label{fig:mountain-car-trees}
	\end{minipage}
\end{figure}

\begin{figure}
	\centering
	\begin{minipage}{0.45\linewidth}
		\centering
		\includegraphics[width=0.95\linewidth]{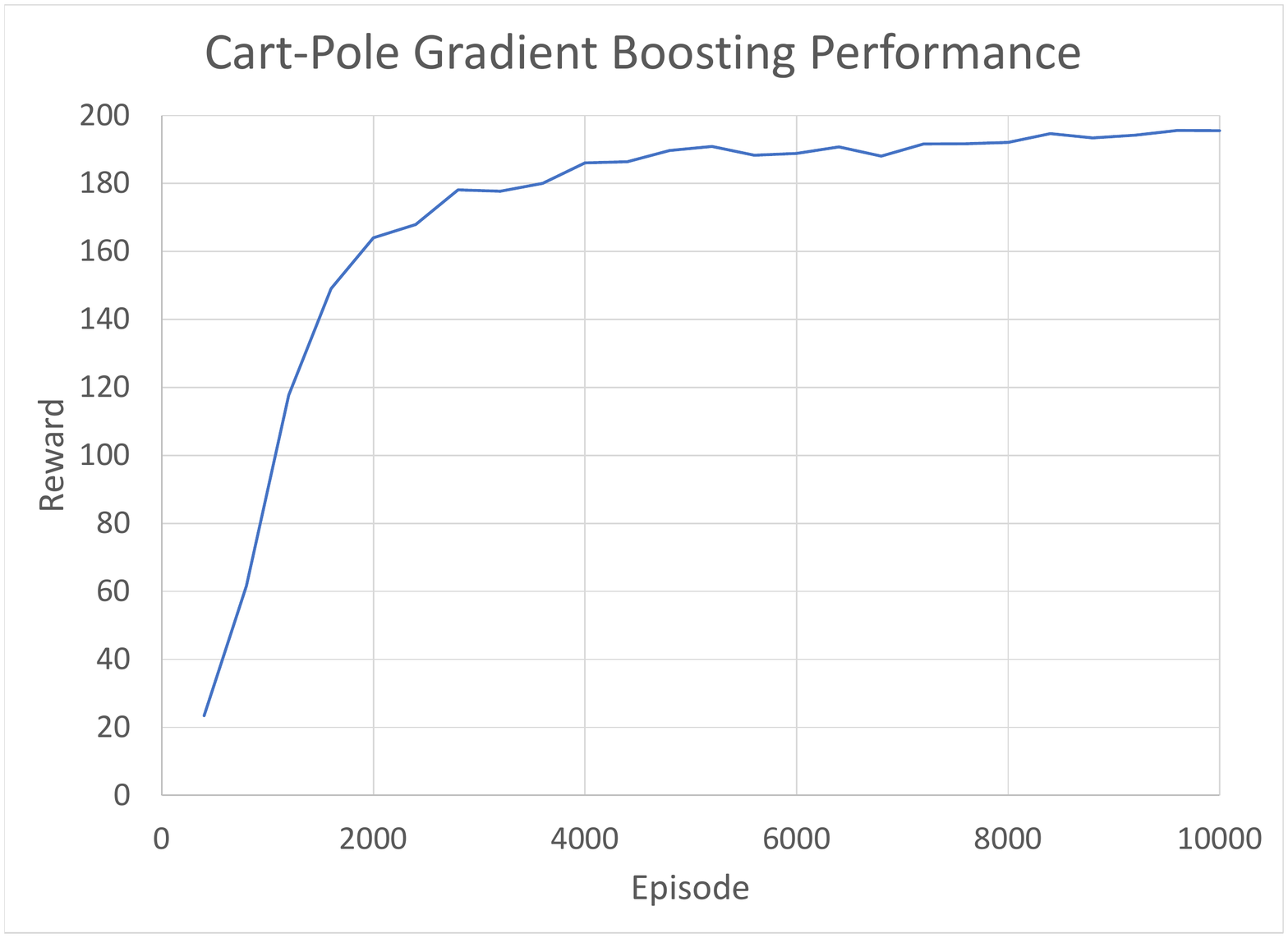}
		\caption{Policy gradient boosting in the cart-pole environment}
		\label{fig:cart-pole-gradient-boosting}
	\end{minipage}%
	\hfill
	\begin{minipage}{0.45\linewidth}
		\centering
		\includegraphics[width=0.95\linewidth]{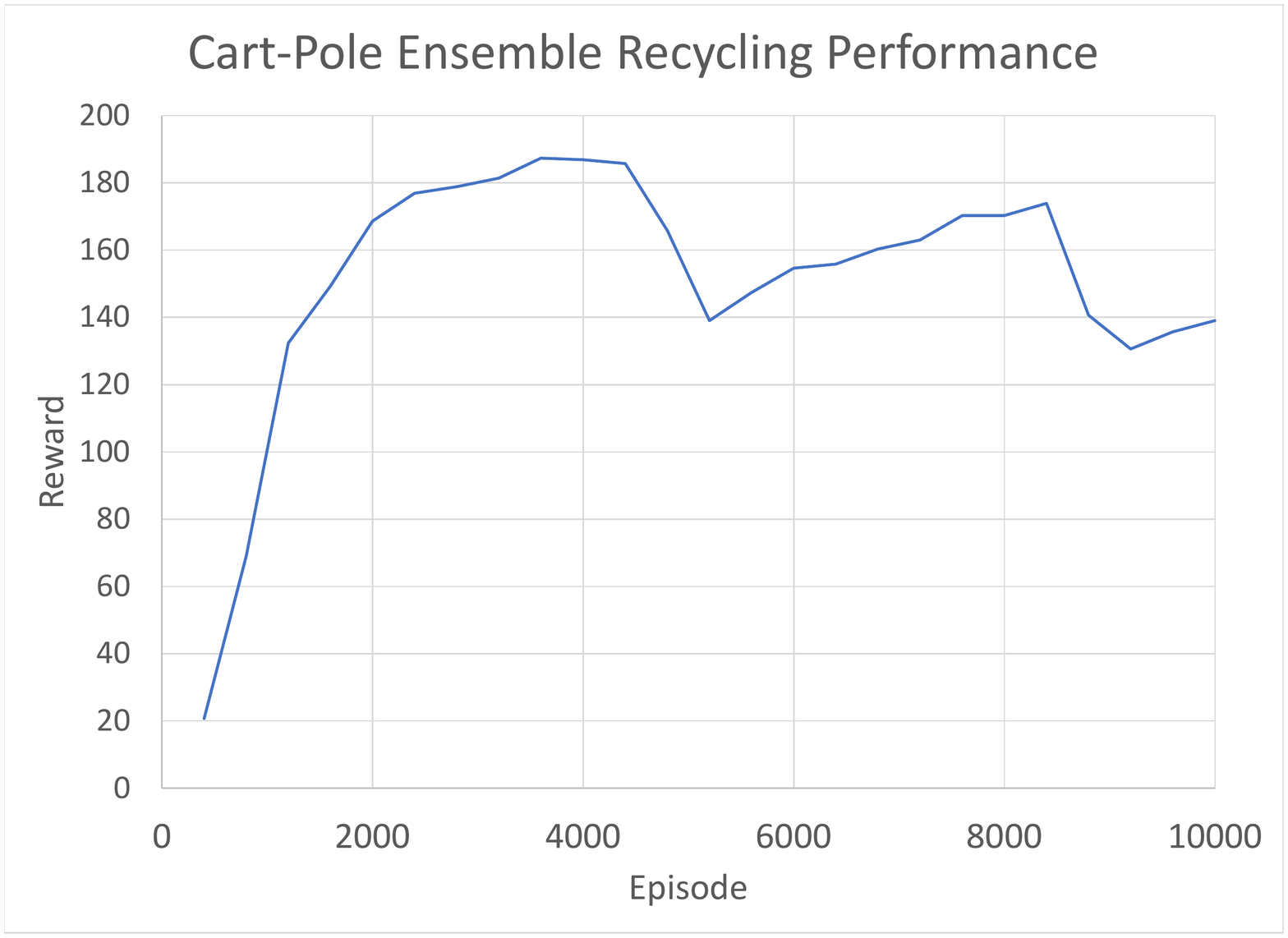}
		\caption{Policy gradient boosting with ensemble recycling in the cart-pole environment}
		\label{fig:ensemble-recycling}
	\end{minipage}
\end{figure}

Policy gradient boosting was successful in the sense that the ensembles did learn how to better behave in the environment. As can be seen in Figure \ref{fig:cart-pole-gradient-boosting}, they learn at a decaying rate approaching the maximum reward per episode of 200. The rate at which they learned, however, was much lower than that of the neural network. The neural network began receiving the maximum reward around episode 400, whereas the ensembles only got close after 8,000 - 10,000 episodes. 

Ensemble recycling was less successful, but does appear to learn in Figure \ref{fig:ensemble-recycling}. Once ensembles begin to be recycled after episode 4,000 the performance starts to dip, and never really recovers. At this stage it does not seem to be a viable alternative to the non-recycling version.

\section{Discussion and Further Research}

Supervised learning on policy data was very successful, to the extent it was tested. The next step for further research is clear: attempt it in more complicated domains. When considering these domains, in particular those with high dimensional action spaces, the question of whether it makes sense to try to learn how to behave with an ensemble of small trees. There has been work to suggest that some such problems may have simple solutions \cite{simsek16}, but it may be challenging for trees to learn those solutions from policy data alone. 

Policy gradient boosting was able to learn how to increase the reward in the cart pole problem, but not to the level we expected. The benchmark is considered to be very easy, yet the ensemble was not able to consistently obtain the maximum reward after thousands of episodes of training. While this result is disappointing, we believe there may still be other feasible ways of incrementally building an interpretable solution to an MDP. Adapting such a solution to conserve space as we attempted with ensemble recycling also still seems to be a reasonable idea, especially in situations where some accuracy can be sacrificed for lower space requirements.

\section{Conclusions}

In conclusion, an ensemble of decision trees can be trained to emulate a policy in simple MDPs. The ensemble is resistant to overfitting and enables humans to better interpret the policy. Further work is necessary to determine if this technique is viable in more complicated domains. An ensemble can also be trained incrementally to perform well in an MDP by policy gradient boosting. This has limitations in that the ensemble can grow to unwieldy sizes before satisfiable performance is achieved, and that the learning process is far slower than modern reinforcement learning methods. Attempting to recycle trees in a simple way in order to limit the size of the ensemble reduces performance in the MDP so much that training should be halted before recycling begins for peak reward.

\section{Acknowledgments}

We would like to thank the Hamel Center for Undergraduate Research for facilitating the SURF program, and Mr. Dana Hamel and Brad Larsen for their generosity.

\nocite{Abel2016}
\nocite{Hastie2009}
\nocite{Petrik2016}

\printbibliography

\end{document}